\pdfoutput=1
\documentclass[11pt]{article}

\usepackage[]{ACL2023}
\usepackage{enumerate}
\usepackage{enumitem}
\usepackage{paralist}
\usepackage{booktabs}
\usepackage{subcaption}
\usepackage{times}
\usepackage{latexsym}
\usepackage{multirow}
\usepackage{url} 
\usepackage{subcaption}
\usepackage{bbding}
\usepackage{amsmath}
\usepackage{threeparttable}
\usepackage{geometry}
\usepackage[T1]{fontenc}
\usepackage[utf8]{inputenc}

\usepackage{microtype}

\usepackage{inconsolata}

\usepackage{graphicx}
\usepackage{booktabs}
\usepackage{graphicx}
\title{A Statistical and Multi-Perspective Revisiting of the Membership Inference Attack in Large Language Models}

\author{Bowen Chen, Namgi Han, Yusuke Miyao \\
  Department of Computer Science,  The University of Tokyo \\
  \texttt{\{bwchen, hng88, yusuke\}@is.s.u-tokyo.ac.jp} \\
}

\begin{document}
\maketitle
\begin{abstract}
The lack of data transparency in Large Language Models (LLMs) has highlighted the importance of Membership Inference Attack (MIA), which differentiates trained (member) and untrained (non-member) data.
Though it shows success in previous studies, recent research reported a near-random performance in different settings, highlighting a significant performance inconsistency.
We assume that a single setting doesn't represent the distribution of the vast corpora, causing members and non-members with different distributions to be sampled and causing inconsistency.
In this study, instead of a single setting, we statistically revisit MIA methods from various settings with thousands of experiments for each MIA method, along with study in text feature, embedding, threshold decision, and decoding dynamics of members and non-members.
We found that (1) MIA performance improves with model size and varies with domains, while most methods do not statistically outperform baselines, (2) Though MIA  performance is generally low, a notable amount of differentiable member and non-member outliers exists and vary across MIA methods,
(3) Deciding a threshold to separate members and non-members is an overlooked challenge, (4) Text dissimilarity and long text benefit MIA performance, (5) Differentiable or not is reflected in the LLM embedding, (6) Member and non-members show different decoding dynamics.
\end{abstract}

\section{Introduction}

Large Language Models (LLMs) \cite{minaee2024largelanguagemodelssurvey} are trained with terabyte level corpora   \cite{chowdhery2022palmscalinglanguagemodeling} that are automatically collected, even the data creators themselves can hardly give instance-level analysis over the collected corpora \cite{biderman2022datasheetpile}.
Such a situation has led to several issues, such as the data leakage of evaluation benchmarks \cite{sainz-etal-2023-nlp} and personal information \cite{YAO2024100211}.

Those concerns inspired the research of the Membership Inference Attacks (MIA) in LLMs \cite{hu2022membershipinferenceattacksmachine}.
\begin{figure}[t] 
\centering % 让图像居中
\includegraphics[width=0.85\columnwidth, height=0.7\columnwidth]{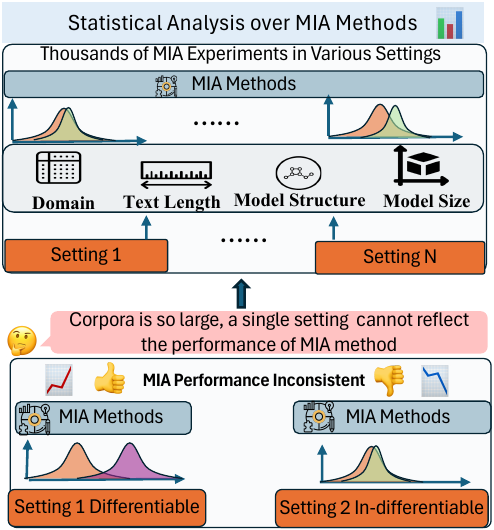} 
\caption{Sample with different settings may result in MIA performance inconsistency.}
\label{fig: research scope}
\end{figure}
Given a set of examples, MIA focuses on differentiating members (trained) and non-members (untrained) by calculating a feature value for each example and splitting them using a threshold.
Generally, those methods focus on observing the outputs of LLMs like generated tokens, probability distributions, losses, etc., and utilize such features to distinguish between members and non-members.
Despite their success in previous studies, recent studies have shown that those methods behave nearly randomly in another MIA construction setting, or such benchmarks can be easily cheated \cite{duan2024membershipinferenceattackswork, das2024blindbaselinesbeatmembership}.
Those negative results raised an inconsistency regarding the performance of MIA methods, e.g., \textit{do those MIA methods really work or not ?}

We see such inconsistency comes from the distribution of the enormous size of the pre-train corpora, which is possible that members and non-members sampled from one setting could be totally different from another setting, leading to inconsistent results.
In this study, instead of a single setting, we evaluate MIA methods statistically from multiple perspectives, e.g., the split methods, domains, text length, and model sizes.
This led to thousands of MIA experiments for one MIA method and enabled a statistical analysis for MIA methods.
Additionally, we conducted an in-depth analysis to study how the text feature, embedding, threshold decision, and decoding dynamics in members and non-members relate to MIA. We found that:
\begin{asparaenum}[(I)]
\item MIA performance improves with model size and varies with domains, while most methods do not statistically outperform baselines.
\item While MIA performance is generally low, a notable amount of differentiable member and non-member outliers exist and vary across MIA methods, connecting the inconsistency regarding the MIA performance.
\item The threshold to separate members and non-members changes with model size and domains, raising it as an overlooked challenge when using MIA in real-world. 
\item While the actual relation varies based on MIA methods, MIA performance generally positively relates to text length and text dissimilarity between members and non-members.
\item Whether members and non-members are differentiable is reflected in LLM embedding with an emergent change in a larger model that makes them more separable.
Specifically, the last layer embedding used by the current MIA methods actually has a low embedding separability.
\item Domains with high MIA performance show a faster increase in accumulated entropy difference for members and non-members.

\end{asparaenum}

\section{Related Works}
Membership Inference Attacks (MIA) \cite{hu2022membershipinferenceattacksmachine} differentiates the member (trained) and non-member (untrained) data by calculating feature values and deciding a threshold for classification.
\subsection{Membership Inference Attack Methods}
First, we introduce MIA methods used in LLM based on the model's transparency.
\paragraph{\textbf{Gray-Box Method}} This method requires the intermediate outputs to be observable, like loss, token probability, etc.
\citet{274574} calculated the loss difference with another reference model with the assumption that if two models are trained under two samples of the same distribution, then the loss of non-members should be significantly different.
Mink-$k\%$ \cite{shi2024detectingpretrainingdatalarge} calculates the average log-likelihood of the tokens with the bottom-$k\%$ output probabilities, 
 suggesting non-member text has more outliers and thus higher negative log-likelihood.
 \citet{zhang2024minkimprovedbaselinedetecting} improved Mink-$k\%$  by standardizing with variance and mean.
\citet{zhang-etal-2024-pretraining} compared predicted token probabilities against actual token probabilities from open corpora, in which the member data should have a closer distribution distance.
Additionally, some methods alter the input text, like token swapping \cite{ye-etal-2024-data} or adding prefixes \cite{xie2024ReCaLL} with the hypothesis that the likelihood of member data should be influenced more by such text alternation.
\paragraph{\textbf{Black-Box Method}} This method only observes the output tokens from the LLM.
\citet{dong2024generalizationmemorizationdatacontamination} calculated a variant of edit distance with multiple generations from the LLM with the hypothesis that those generations of a member text should have a smaller lexical distance compared to non-member text.
Additionally, \citet{kaneko2024samplingbasedpseudolikelihoodmembershipinference} made a similar hypothesis while they evaluated the semantic similarity using the embedding model.

% \subsection{Membership Inference Attack Methods}
%  researchers utilize the loss \cite{8429311} and its Zlib entropy \cite{gailly2004zlib} calibrated version, which considers both the number of bits required to compress the given text and the corresponding loss of this instance.
% Recently, LLM-specified gray-box MIA attacks have been developed.

\subsection{Membership Inference Attack Analysis}
Regarding the MIA analysis, some research \cite{maini2021datasetinferenceownershipresolution,carlini2022membershipinferenceattacksprinciples} suggest the MIA difficulty increases with model size and corpora size.
\citet{zhang2024membershipinferenceattacksprove} showed with toy data that it is hard to reliably operate an MIA method under a certain false positive rate.
\citet{meeus2024sokmembershipinferenceattacks} found some MIA benchmarks are flawed, which can be easily cheated by just checking the word differences \cite{das2024blindbaselinesbeatmembership}.
\citet{duan2024membershipinferenceattackswork} evaluated Gray-Box MIA methods in the train and test set of pre-train corpora of an LLM, where they behave almost randomly.

Those negative findings show an inconsistency with the 
performance reported by the previous MIA methods.
We assume such inconsistency comes from the sampled member and non-member distribution under different settings, which could be totally different due to the enormous size of corpora, leading to this inconsistency.
In this study, instead of using a single setting, we create various settings, leading to thousands of experiments for one MIA method.
This allows a statistical-level analysis of MIA methods from multiple perspectives, which shows new findings and connects the MIA performance inconsistency.

\section{Experiments Setting}
%\subsection{Problem Statement}
Given a model $M$ and set of data $X = \{x_0 \dots x_n\}$, where each $x$ is a text consisted of $\{t_0 \dots t_m\}$ tokens
a MIA method calculates feature scores $S = f(M; X)  = \{s_0 \dots s_n\}$ for every data instance.
A threshold $t$ will be selected to classify whether $x_i$ belongs to training data $D$ of the model $M$.
Data that are in the $D$ ($x_i \in D$) are called member data, otherwise called non-member data.

\subsection{MIA Methods}

%\begin{table*}[htbp]\small
%\begin{tabular}{@{}lll@{}}
%\toprule
%Name            & Category & Calculated Feature \\ \midrule
%Loss            &          &  $ \text{Example Loss}$      \\
%Perplexity     &          &   $\text{Example Perplexity}$     \\
%Reference Model &          &  $ \text{Loss}_{\text{Target}} - \text{Loss}_{\text{Reference}}$      \\
%Zlib Entropy    &          &   $ \frac{\text{Loss}}{\text{Entropy}(\text{Text})}$     \\
%Min-k\% Prob    &          &  $ \frac{1}{N} \sum_{i=1}^{N} \log P(w_i | w_{<i})$ for $w_i \in \text{Bottom-K\%}$      \\
%Min-k\% Prob ++ &          &   $ \frac{\text{Min-k\% Prob} - \mu}{\sigma}$     \\
%ReCaLL          &          &        \\
%SaMIA           &          &   $\frac{1}{N} \sum_{i=1}^{N} \text{SemanticDistance}(g_i, a)$ .     \\
%CDD             &          &   $ \frac{1}{N} \sum_{i=1}^{N}\text{EditDistance}(g_i, a)$.     %\\ \bottomrule
%\end{tabular}
%\caption{}
%\label{tab:method table}
%\end{table*}

\subsubsection{Baseline Methods}
\paragraph{\textbf{Loss \cite{8429311}}} collects the loss value $L(M_t; x)$ for each input instance text.

\paragraph{\textbf{Refer \cite{274574}}} calculates the loss gap between the attacked model $M_t$ and a reference model $M_r$ for the input text $L(M_t; x)-L(M_r;x)$. 

\paragraph{\textbf{Gradient}} collects the gradient value $G(M_t; x)$ for each input instance text.

\paragraph{\textbf{Zlib \cite{274574}}} calibrates the loss by the Zlib compression entropy $Z(x)$ of the input text, calculated as $\frac{L(M_t; x)}{Z(x)}$.
\subsubsection{Token Distribution Based Method}
Those methods hypothesize that non-member text contains more rare tokens or has a different distribution whose average log-likelihood should be different than that of member text.
\paragraph{\textbf{Min-$k\%$ Prob \cite{shi2024detectingpretrainingdatalarge}}} calculates the average log-likelihood of tokens in bottom-$k\%$ decoding probabilities in the whole input tokens $\mathrm{Bot}(x)$, which is calculated as
$\mathrm{MinK}(M_t; x) = \frac{1}{E}\sum_{t_i \in \mathrm{Bot}(x)} \log p(t_i \mid t_1, \ldots, t_{i-1})$.
$E$ represents the number of bottom-$k\%$ tokens.

\paragraph{\textbf{Min-$k\%$ Prob++ \cite{zhang2024minkimprovedbaselinedetecting}}} standardizes Mink-$k\%$ with its mean and standard deviation.
As the Mink-$k\%$ did not standardize the value, causing an unstable value range.

\paragraph{\textbf{DC-PDD} \cite{zhang-etal-2024-pretraining}} computes the divergence
between the probability of decoded tokens with their token probability distribution pre-computed based on a large corpora.\footnote{We cannot fully reproduce this method as its pre-computed frequency in large corpora is not released.}

\subsubsection{Text Alternation Based}
This method alters the input text by adding a prefix or swapping tokens with the hypothesis that the log-likelihood of member text is affected more than non-member text in such perturbation.

\paragraph{\textbf{EDA-PAC} \cite{ye-etal-2024-data}} creates a perturbed text $\hat{x}$ by continuously swapping two random tokens.
Then, it calculates the difference between the average log-likelihood for Top-$k\%$ and Bottom-$k\%$ tokens for $x$ and the swapped $\hat{x}$.
The hypothesis is that the token swap alters a member to a non-member while a non-member is still a non-member, so the members should be influenced more.

\paragraph{\textbf{\textsc
{ReCaLL} \cite{xie2024ReCaLL}}} creates a non-member prefix $p$ and concate it with text $x$ to calculate a \textsc{ReCaLL} score $\frac{LL(x|p)}{LL(x)}$ where $LL$ is the average log-likelihood.
The hypothesis is that if $x$ is a member text, the non-member prefix $p$ perturbs LLM's confidence in generating it, while such perturbation affects less for non-members.

\subsubsection{Black-Box Methods}
This method generates multiple continuations for a text prefix with the hypothesis that the multiple generated continuations of member text should have a higher semantic/lexical similarity with the actual continuations than those of non-member text.
\paragraph{\textbf{SaMIA \cite{kaneko2024samplingbasedpseudolikelihoodmembershipinference}}} inputs a partial prefix of the text and generates multiple continuations. 
Then, it calculates the average semantic similarity of the generated continuations with the actual continuations as the feature value.

\paragraph{\textbf{CDD \cite{dong2024generalizationmemorizationdatacontamination}}} inputs a prefix of the text and generates multiple continuations for this prefix.
Then, it calculates a variant of edit distance between generated continuations with the actual continuation to calculate the peakiness score, e.g., a measurement of how generated tokens are similar to each other on the token level.
\footnote{Detaield experiment settings are in Appendix \ref{sec: mia setting}}

\subsection{Datasets}
We use one existing benchmark and sample data from pre-train corpora with various settings.
\paragraph{\textbf{WikiMIA \cite{shi2023detecting}}} contains Wikipedia text sampled at the timestamp of 2023/10.
Text samples before the time stamp are member text, and those after them are non-member text.
This benchmark has been used by several MIA methods \cite{kaneko2024samplingbasedpseudolikelihoodmembershipinference, zhang-etal-2024-pretraining}, which shows that its member and non-member splits are separable.
Thus, we use it as a separable MIA benchmark in some experiments for reference.

% \paragraph{\textbf{TemporalWiki \cite{duan2024membershipinferenceattackswork}}} contains Arxiv abstract sampled from different timestamp [2020/08, 2021/01, 2021/06, 2022/01, 2022/06, 2023/01, 2023/06] where the later timestamp can be used as non-member text and previous timestamp can be used as member text.

\paragraph{\textbf{Pile \cite{gao2020pile800gbdatasetdiverse}}} corpora contains texts collected from domains like arXiv, GitHub, Freelaw, PubMed, DM Math, etc, with the train, test, and validation sets.
Train set text and test set text can be treated as member text and non-member text, which will be the main focus of this study.

\subsection{MIA Data Construction for Pile}
We provide three split methods to construct the member and non-members set for the Pile dataset.

\paragraph{\textbf{Truncate Split \cite{duan2024membershipinferenceattackswork}}} creates the member set and non-member set by truncating texts into a fixed range.
We extend it by setting a length range of 100 from 0 to 1000. 

\paragraph{\textbf{Complete Split}} samples member and non-member texts whose whole length is in a text range that follows the Truncate Split.

\paragraph{\textbf{Relative Split}}  calculates the ten-percental text length range based on the test set of each domain.
The member and non-member text are sampled from those ten-percentile length ranges.

Each split method is applied to all domains in the Pile, with a minimum of 100 examples for both members and non-members. 
As text distribution varies by domain, not every domain meets this criterion. 
\footnote{For example, two split methods may keep 8 and 5 out of 10 domains. Domains in each split are in Appendix Table \ref{tab:all_domains}.}
This resulted in nearly 100 GBs of member and non-member texts sampled from different settings for MIA experiments.

This statistical evaluation contains (1) more domain coverage (compared to WikiMIA, ArxivMIA \cite{shi2023detecting}, BookMIA \cite{shi2024detectingpretrainingdatalarge}), (2) broader text length range (compared to MIMIR \cite{duan2024membershipinferenceattackswork}),
(3) considered the truncation method and domain-specific sampling.
We run all MIA methods on every length split on domains in that split for every model size in all random seeds with 4,860 experiments for one MIA method.
%\footnote{For example, running one MIA method on all model sizes, domains, split methods,  text lengths, and random seeds will result in 4,860 MIA experiments.}

% Those construction methods are applied to all domains with 1,000 maximum samples for both member and non-member text and discard domains that do not have 100 samples.
% As different domains have different text distributions, not all domains are shared across every split method.

% This results in experiment results of $m \times length \times domains \times model\ size$. 
\subsection{Models}
We use the Pythia model \cite{biderman2023pythiasuiteanalyzinglarge} (160m, 410m, 1b, 2.8b, 6.9b, 12b) that trained on the deduplicated Pile corpora to avoid effects from duplicate texts.
It contains train, valid, and test sets.
The valid and test set texts are treated as non-member texts.
%\footnote{We use the Pythia model trained on the deduplicated Pile corpora to avoid the effect brought by duplicate texts.}
%\footnote{Regarding model size, we use $m$ to represent million and $b$ to represent billion in following sections.}

\subsection{Evaluation Metric}
\paragraph{\textbf{ROC-AUC \cite{FAWCETT2006861}}} iterates every threshold for binary classification to calculate the Ture Positive Rate (TPR) and False Positive Rate (FPR) to form a ROC curve. AUC is the area under this curve and is used to analyze MIA performance.

%\paragraph{\textbf{Kolmogorov-Smirnov Test (K-S Test) \cite{inbook}}} verifies whether two given distributions are drawn from the same distribution through a hypothesis test.
%If the result K-S Test P-value is below 0.05, it means those two samples are not drawn from the same distribution. 

% \paragraph{\textbf{Jason Shannon Distance (J-S Distance) \cite{10.1162/coli.2000.26.2.277}}}  measures the similarity of two distribution from a information theory perspective based on entropy. 
% It means how much new information can be obtained from the the target distributino from the persepctive of evaluated distribution.
% \paragraph{\textbf{Wasserstein Distance \cite{ramdas2015wassersteinsampletestingrelated}}}  measures the mimum cost of turning one distribution into another distribution.

\paragraph{\textbf{Davies-Bouldin Score \cite{shi2023detecting}} (DB-Index)} evaluates the separability of two clusters of embeddings.
A lower value indicates a better seperability.
This is used to evaluate the separability of embeddings for members and non-members. 

\begin{figure*}[t] 
\centering % 让图像居中
\includegraphics[width=1\textwidth]{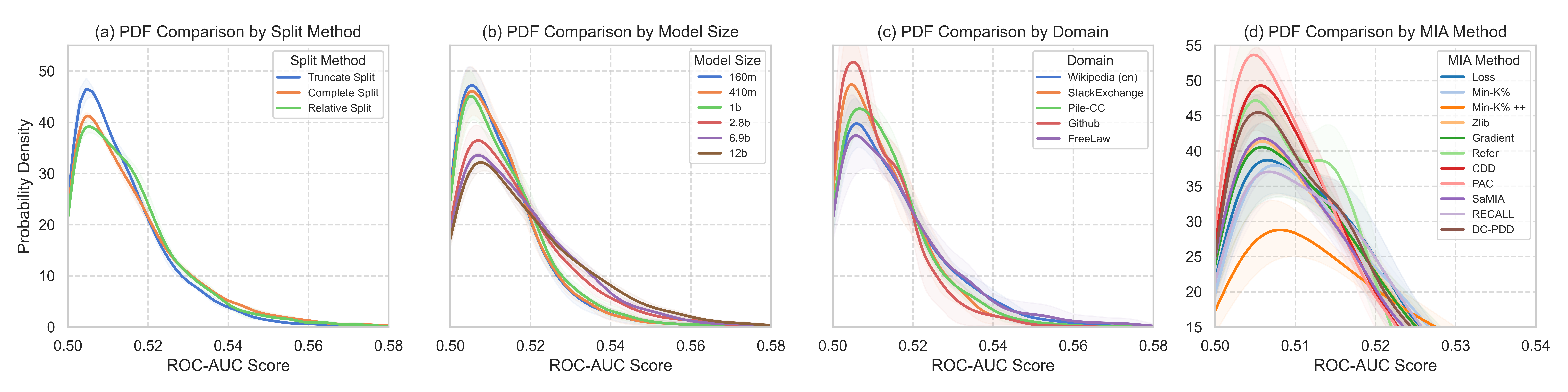} 
\caption{ROC-AUC probability density in different dimensions while fixing other dimensions. Less area on the left side means statistically better MIA performance. Shade area means variance from random seeds. We only enlarge Figure (d) to increase the readability due to the number of MIA methods.}
\label{fig: general results}
\end{figure*}
\section{Results}
We first statistically analyze the ROC-AUC scores of MIA methods, which generally align with previous negative results but also show new findings.
Then, we analyze the outliers where the members and non-members show differentiability and connect the inconsistency regarding MIA performance.
Next, we discuss the threshold decision when using the MIA  to analyze its real-world effectiveness. 
Additionally, we explore how MIA is related to the input text itself by studying its correlation with text length and similarity.
Finally, we seek the explanation for MIA performance from the LLM structure level with an analysis of the separability of embeddings and decoding dynamics for members and non-members.
\subsection{Effect of Different Factors}
\label{sec}
We aggregate ROC-AUC scores between 0.50 and 0.58, cover most of the experiments, and calculate their probability density over the split method, model size, domain, or MIA methods while fixing the others in Figure \ref{fig: general results}.
While 0.50-0.52 occupies most probability densities, we still observe that: 
\begin{asparaenum}[(I)]
\item  In Figure 2 (a), the commonly used Truncate Split shows the worst performance, while the Relative Split gives the best performance.
Truncating a text may cause it to lose outlier words.
Additionally, such loss of contextual information affects MIA methods that rely on alternating the original members and also affects the Black-Box method as the quality of generated tokens deteriorates.
\item In Figure 2 (b), MIA performance improves with model size, particularly from 1b to 2.8b, which contradicts previous findings that suggest it should decrease with model size.
We think that a small model struggles with learning large corpora due to a small capacity, causing most member texts to behave like non-members and reducing MIA performance.
As model capacity scales, more member texts are well learned, which starts to differ from non-member text and enhances performance. 
%However, this does falsify previous research.
Our results do not falsify previous research.
If a much larger LLM learns very well and even fits well with non-member text, it may again show a low MIA performance.
Thus, the model size and MIA performance relation may be an inverse U-curve.
\item In Figure 2 (c), among shared domains across split methods, Wikipedia (en) and FreeLaw show statistically better performance compared to other domains.
We suggest this is related to token diversity.
GitHub and StackExchange are related to codes that have less token diversity compared to FreeLaw and Wikipedia, where various words are used.
The Pile-CC is a general domain that contains various texts whose token diversity is between the text domain and code domain.\footnote{ Appendix \ref{other pdfs} contains results for all domains in each split method.}
\item  In Figure 2 (d), only PAC and CDD are worse than the Refer baseline, and the Loss baseline is only 
 outperformed by Min-k\% ++, Min-k\%, and \textsc{ReCaLL}.
Other methods are between those baselines, and their performance gap is within the variance from random seeds.
However, this does not indicate their peak performance in certain settings since the probability density tests the generalizability of the hypothesis in each MIA method.
\end{asparaenum}

\subsection{Outliers in MIA}
While the MIA performance is generally low, we still observed notable outliers with relatively high differentiability (ROC-AUC > 0.55) not captured by the probability density.
\begin{table}[t]\scriptsize
\centering
\setlength{\tabcolsep}{3pt}
\begingroup
\begin{tabular}{lccccccccc}
\toprule
\multirow{2}{*}{\textbf{Method}}& \multicolumn{7}{c}{\textbf{Model Size}}& \multicolumn{2}{c}{\textbf{ROC-AUC}}\\
\cmidrule(lr){2-8}\cmidrule(lr){9-10}
 &  160m &  410m &  1b &  2.8b &  6.9b &  12b &  Num &  Max &  Mean\\
\midrule
Loss       &    13 &    12 &  12 &    20 &    23 &   30 &             110 &    .585 &     .561 \\
Gradient   &     7 &    14 &   8 &    54 &    46 &   31 &             160 &    .631 &     .563 \\
Refer      &    12 &    12 &  12 &    12 &    11 &   11 &              70 &    .572 &     .559 \\
Zlib       &    12 &    12 &  13 &    22 &    24 &   47 &             130 &    .590 &     .562 \\
Min-K\%     &     8 &    12 &  11 &    25 &    23 &   48 &             127 &    .600 &     .562 \\
Min-K\% ++  &     8 &    11 &  17 &    \underline{94} &  \underline{107} &  \underline{173} &             \underline{410} &    .631 &     .564 \\
DC-PDD     &     0 &     0 &   0 &    12 &    16 &   15 &              43 &    .575 &     .558 \\
PAC        &     5 &     4 &   2 &     1 &     3 &    5 &              20 &    .573 &     .557 \\
ReCaLL     &    14 &    15 &  16 &    24 &    25 &   33 &             127 &    \underline{.806} &     \underline{.572} \\
SaMIA      &    \underline{40} &    \underline{37} &  \underline{38} &    37 &    32 &   34 &             218 &    .647 &     .569 \\
CDD        &     7 &    17 &  11 &    12 &    10 &    6 &              63 &    .604 &     .561 \\
\bottomrule
\end{tabular}
\endgroup
\caption{The number of differentiable outliers across model size and MIA methods along with max and mean ROC-AUC scores. Underscored mean highest value.}
\label{tab: outliers}
\end{table}
\subsubsection{Outliers Statistics Analysis}
We count those outliers across the MIA method and model size along with their maximum and mean ROC-AUC values in Table \ref{tab: outliers}.
\begin{figure*}[htbp] 
\centering
\includegraphics[width=1\textwidth, height=0.35\textwidth]{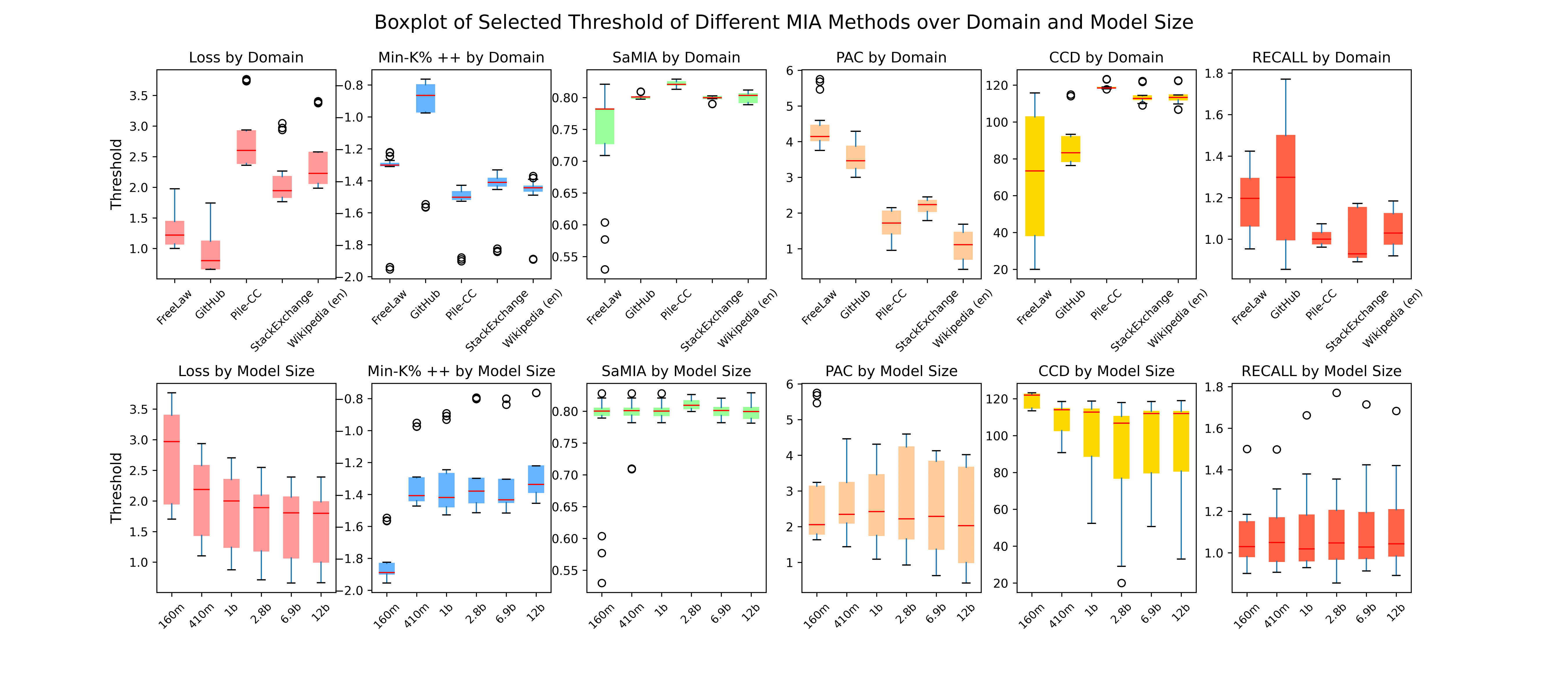} 
\caption{Boxplot of the threshold for different MIA methods over domains and model sizes.}
\label{fig: threshold_box}
\end{figure*}
\begin{figure}[htbp] 
\centering
\includegraphics[width=1\columnwidth, height=0.65\columnwidth]{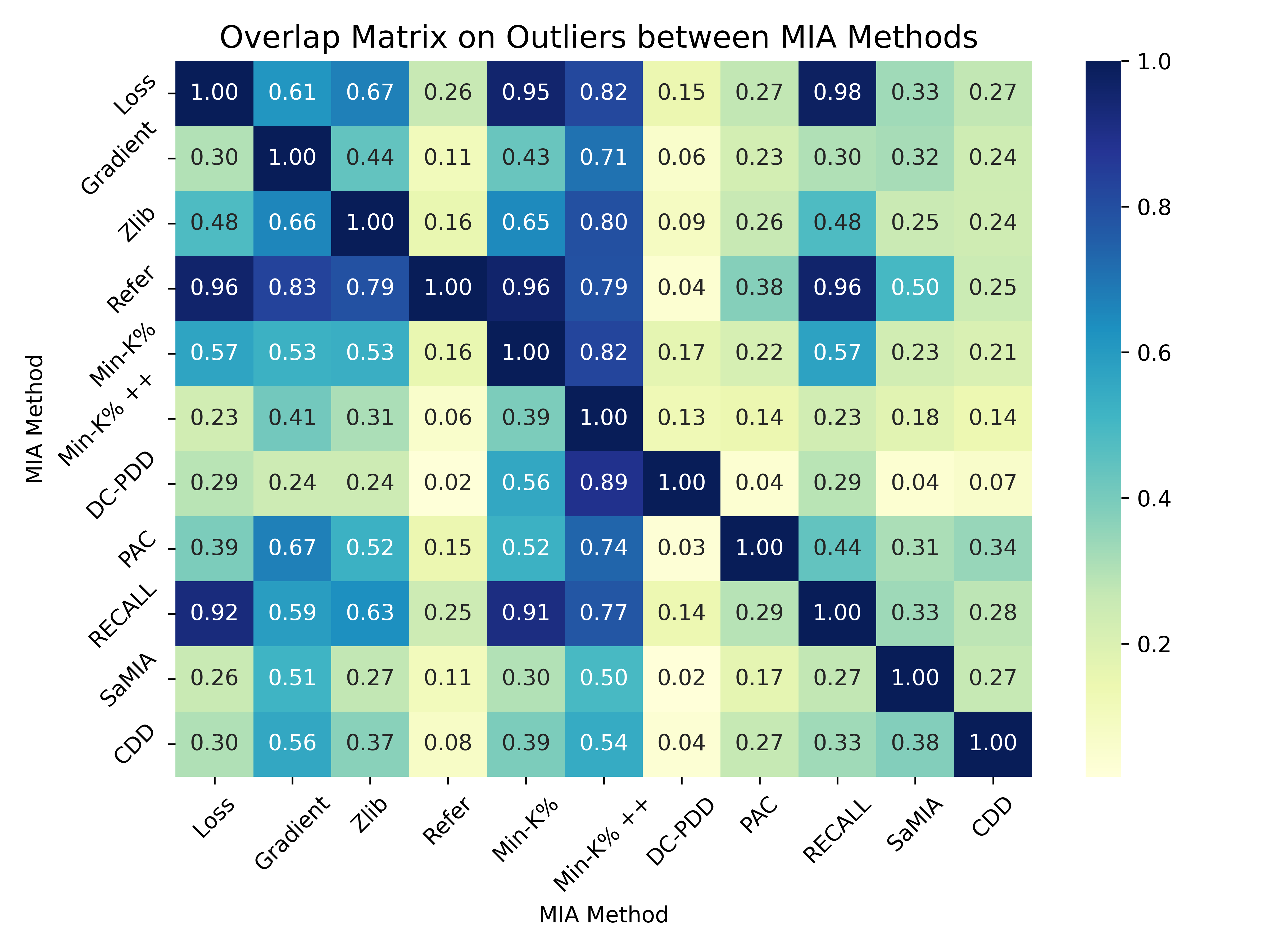} 
\caption{MIA outliers overlap matrix across methods.}
\label{fig: overlap}
\end{figure}
\begin{asparaenum}[(I)] 
\item Those outliers occupy a small ratio with 8.4\% even for Min-k\% ++, which generally aligns with previous negative results regarding the MIA performance.
However, the existence of those outliers also provides space for previous positive results, connecting their inconsistency.
\item In most MIA methods, the number of differentiable splits increases with model size.
As results in the section \ref{sec} already show, large models statistically perform better. 
We hypothesize that the internal structure of LLM changed in a way that positively affects MIA when scaling model size.
However, methods (SaMIA, CDD, Refer) that do not only rely on internal states of LLM are less sensitive to increasing model size.
\item Additionally, we also see that the maximum and mean performance are not related to how many outliers exist in the MIA methods.
The highest value reaches 0.81 in \textsc{ReCaLL} while the number of its outliers is not either the highest or lowest.
This suggests the method that works generally better (Min-K\% ++) does not mean it is also the absolute better one, supporting that a MIA method should be evaluated statistically.
\end{asparaenum}

\subsubsection{MIA Methods Consistency on Outliers}

With the existence of outliers, we study whether they are consistent across MIA methods by calculating their overlap ratio in Figure \ref{fig: overlap}.
\begin{asparaenum}[(I)] 
\item Even the best-performed method (Min-$k\% $++) does not have a general higher overlap, which only has a 4\% overlap with the CDD method.
SaMIA and CDD give a low overlap when compared to all other MIA methods as they do not require any internal outputs, which is significantly different from other MIA methods.
\item Even though most methods do not statistically outperform baselines, this does not mean those methods are not meaningful, as the overlap matrix shows each MIA method works in different situations.
The results also suggest it is hard to use one hypothesis to outperform all others.
\end{asparaenum}

\subsection{Generalization of Threshold in MIA}
The ROC-AUC metric iterates feature values to differentiate between members and non-members but does not show how to decide a threshold and its general effectiveness in MIA. 
To address this, we split the member and non-member sets into training and validation sets in a 4:1 ratio and use the Geometric Mean $t = \arg\max_{i} \sqrt{TPR_i \times (1 - FPR_i)}$ \cite{youden1950index} to find a threshold that balances the true positive rate and false positive rate. 
The distribution of this threshold across different model sizes and domains is shown in Figure \ref{fig: threshold_box}.\footnote{Boxplot of other methods are in Appendix \ref{other boxplots}.}
\begin{table*}[ht]
\centering
\setlength{\tabcolsep}{4pt}
\scriptsize
\begin{tabular}{lcccccccccccccccccccc}
\toprule
\multirow{3}{*}{\textbf{Method}} &\multicolumn{10}{c}{\textbf{Text Length}} & \multicolumn{10}{c}{\textbf{Text Similarity}}\\
\cmidrule(lr){2-11}
\cmidrule(lr){12-21}
& 
\multicolumn{3}{c}{\textbf{Truncated}} & \multicolumn{3}{c}{\textbf{Complete}} & \multicolumn{3}{c}{\textbf{Relative}} & \multirow{2}{*}{\textbf{Avg}}&\multicolumn{3}{c}{\textbf{Truncated}} & \multicolumn{3}{c}{\textbf{Complete}} & \multicolumn{3}{c}{\textbf{Relative}} & \multirow{2}{*}{\textbf{Avg}}  \\
\cmidrule(lr){2-4}\cmidrule(lr){5-7}\cmidrule(lr){8-10}
\cmidrule(lr){12-14}\cmidrule(lr){15-17}\cmidrule(lr){18-20}
 & FL & Pi & Gi & FL & Pi & Gi& FL & Pi & Gi& &FL & Pi & Gi & FL & Pi & Gi& FL & Pi & Gi \\
\midrule
Loss & \underline{.21} & .11 & .05 & .21 & .19 & \underline{.28}& \underline{\textbf{.29}} & .18 & .17 & .17&.04&\underline{-.17}&-.09&\underline{-.21}&-.03&-.11&-.21&\underline{\textbf{-.24}}&-.14&-.13\\
Refer & -.19 & -.09 & \underline{.31} & \underline{.21} & -.18 & -.35& .24 & .28 & \underline{\textbf{.34}} &.11 &\underline{-.35}&-.01&-.25&.18&.29&\underline{-.32}&-.30&-.30&\underline{\textbf{-.40}}&-.16\\
Zlib & \underline{.32} & .06 & .00 & .32 & \underline{.38} & .36 & \underline{\textbf{.39}} & .21 & .05 &.23&\underline{-.22}&-.22&-.04&-.21&-.26&\underline{\textbf{-.40}}&-.21&-\underline{.25}&-.01&-.20\\
Min-k\% & \underline{.42} & .21 & .28 & .36 & .22 & .15 & \underline{\textbf{.47}} & .42 & .13 &.29&\underline{-.30}&-.28&-.06&-.13&\underline{-.21}&-.16&-.29&-.06&\underline{\textbf{-.32}}&-.20\\
Min-k\%++ & \underline{\textbf{.81}} & .56 & .33 & \underline{.78} & .73 & .42& .48 & \underline{.76} & .21 & \textbf{.56} &\underline{\textbf{-.75}}&-.58&-.16&\underline{-.71}&-.62&-.26&-.56&\underline{-.60}&-.12&\textbf{-.48}\\
DC-PDD& .48 & .34  & .02 & .12 & .19 & .10& .25 & .08 & .34 &.21&\underline{\textbf{-.46}}&-.34&-.01&-.11&\underline{-.23}& -.09&-.26&-.12&\underline{-.32}& -.22\\
PAC & -.07 & .12 & \underline{.16} & \underline{.21} & -.03 & .15 &\underline{\textbf{.25}} & -.18 & -.03& .06 &.02&-.14&\underline{-.15}&-.16&.15&\underline{\textbf{-.62}}&-.05&\underline{-.17}&-.04&-.14 \\
\textsc{ReCaLL} & .11 & .15&\underline{.25} & .21 & .19 & \underline{\textbf{.29}}& \underline{.19} & .13 & .15& .21 &.04&\underline{-.13}&-.09&-.03&-.03&\underline{-.10}&\underline{\textbf{-.20}}&-.08&-.08&-.08\\
SaMIA & -.34 & \underline{.02} & -.14 & -.12 & -.22 & -.50& \underline{-.08} & -.05 & -.26& -.18 &-\underline{\textbf{.31}}&-.08&-.13&\underline{-.20}&-.04&-.17&-.03&-.17&\underline{-.20}&-.14\\
CDD & .18 & \underline{.30} & -.03 & \underline{-.11} & -.26 & -.11& \underline{\textbf{.44}} & .24 & -.42 &.03&.10&\underline{-.29}&.05&-.02&.11&\underline{-.03}&\underline{\textbf{-.55}}&-.27&-.42& -.15\\
\hline
Avg & .16 & .16&\underline{.17}&\underline{.17}&0.11&.15&\underline{\textbf{.28}}&.22&.02&.16&\underline{-.22}&\underline{-.22}&-.09&-.16&-.08&\underline{-.23}&\underline{\textbf{-.26}}&-.23&-.21&-.19 \\
\bottomrule
\end{tabular}
\caption{Spearman coefficient between RUC-AOC with Text Length and Similarity across MIA methods in FreeLaw (FL), Pile-CC (Pi), and GitHub (Gi) domains. We underscore the highest/lowest value in that split and bold those that are highest among splits, where the highest is used for text length, and the lowest is used for text similarity.}
\label{tab:combined_results}
\end{table*}
\begin{asparaenum}[(I)] 
\item  In the top figure, the threshold varies not just between domains but also within the same domain with the existence of outliers.
In the bottom figure, the threshold changes with model sizes, as most MIA methods rely on the output likelihood, which is related to the model size.
The SaMIA, which relies on an external model to compare sentence similarity, is less affected by the model size, further confirming this point.
This suggests the threshold in one model size may not work for the others. 
\item These results show the generalizability of the MIA threshold as an overlooked challenge.
A threshold may not work even in samples from the same domain, may not transfer to another domain, and may not work in another model size, leading to a high possibility of performance deterioration when using the MIA method in the real world.

\subsection{Text Similarity and Text Length}
% \begin{table}[!tb]\scriptsize
% \centering
% \begingroup
% \begin{tabular}{lcccc}
% \toprule
% Split & Zpif & CHI(\text{\(\chi^2\)}) & DC & FID\\
% \midrule
% Rela Comp & 0.03&0.17 &0.24 & 0.25\\
% Abs Inc & 0.34& 0.06& 0.37& 0.41\\
% Abs Comp & 0.16&0.14 &0.15 & 0.08\\
% \bottomrule
% \end{tabular}
% \endgroup
% \caption{The Correlation Coefficient Between Split Method and Corpus Similarity Metric}
% \label{tab: corporasim}
% \end{table}
% In this section, we discuss how the input text affects the MIA performance by analyzing the text similarity and text length of members and non-members.
Previous studies showed text length \cite{zhang-etal-2024-pretraining} and token differences \cite{duan2024membershipinferenceattackswork} contribute to the MIA but with results induced from single method or splits, lacking general evidence.
In this section, we calculate the Spearman correlation \cite{spearman} between the ROC-AUC score with text length and the 7-gram overlap occurrence for every MIA split in Table \ref{tab:combined_results}.
\end{asparaenum}

%\subsection{Effect of Text Length}

% In this section, we discuss how the input text length affects the performance of MIA methods.
% We collect the text length that has the best performance for a MIA method on different domains.
% We also apply linear regression to see if the ROC-AUC performance is positively related to the increase in text length, in which we report the slope of the regression line. 
% The results are in Table \ref{tab:combined_results}.

\begin{asparaenum}[(I)] 
\item For most of the MIA methods, its average correlation with length is positive, indicating longer text benefits MIA in general.
However, it also varies based on split methods and MIA methods.
We see that SaMIA and CDD showed a negative and near-zero correlation.
For such Black-Box methods, the generated tokens will largely deviate from the actual continuation for both members and non-members in long text.
The SaMIA used semantic comparison, which is affected more by such a deviation than the lexical distance of CDD.
\item In the text similarity, we see a universal negative relation, indicating that token differences between members and non-members benefit the MIA performance.
However, PAC is less negatively related to text similarity as it inserts a prefix at a running time for both members and non-members, increasing the text similarity.
This also explains one reason for the generally low MIA performance, e.g., they detect word differences rather than member and non-member differences.
\end{asparaenum}
% \begin{figure*}[htbp] 
% \centering
% \includegraphics[width=1\textwidth, height=0.55\textwidth]{figures/all_models_embedding_result.png} 
% \caption{The DB Score and Transformer Classifier Performance (Accuracy) on the Member and Non-Member embeddings. The differentiable domains selected are DM Mathematics, GitHub, and WikiMIA. The indifferentiable domains selected are arXiv, Pile-CC, and PubMed. For each dataset, the dotted line with circle markers means the accuracy, and the solid line with triangle markers means the DB Score.}
% \label{fig: embedding results}
% \end{figure*}
\begin{figure*}[!t] 
\centering
\includegraphics[width=1\textwidth, height=0.55\textwidth]{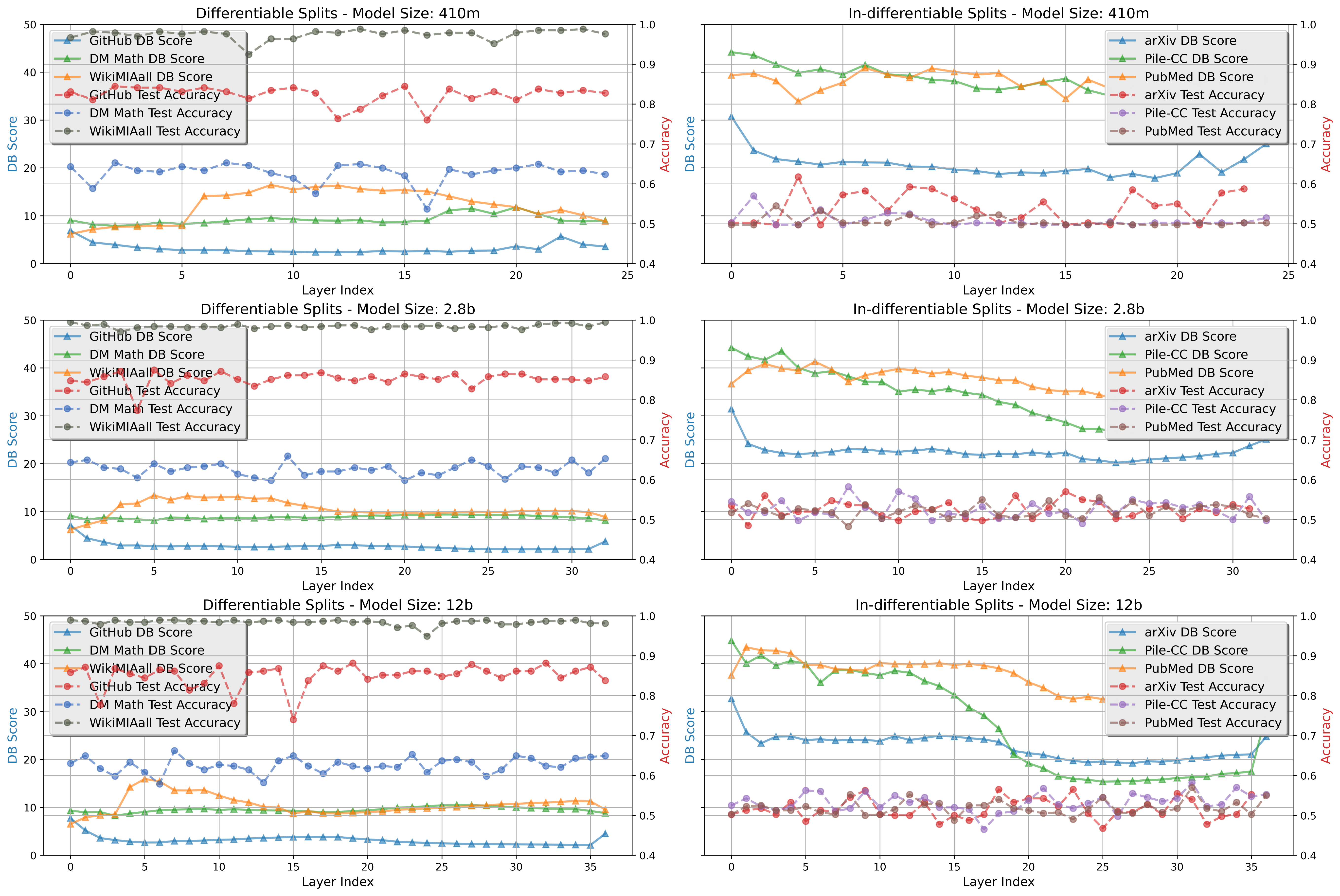} 
\caption{The DB Score (solid line with triangles) and Transformer Classifier Accuracy (dotted line with circles) on the member and non-member embeddings. The differentiable outliers come from DM Math, GitHub, and WikiMIA. The indifferentiable splits come from arXiv, Pile-CC, and PubMed. }
\label{fig: embedding results}
\end{figure*}

\subsection{Embedding Probing and Seperability}
In this section, we discuss how embeddings of members and non-members are represented across layers to answer the question of \textit{are they originally indifferentiable at the internal states?}
We collect the average pooled hidden states at each layer for members and non-members.
The DB Score is used to evaluate how separable those embeddings are, and we train a Transformer classifier on them to see if they are directly separable, as shown in Figure \ref{fig: embedding results}.
\begin{asparaenum}[(I)] 
\item  The DB Score is around 10 in the differentiable splits with 70\%-100\% accuracy in the Transformer classifier and reaches around 40 in the in-differentiable splits with random guess accuracy (50\%).
This suggests differentiable splits are originally easier to differentiate from the embedding level, while their varied accuracies and DB scores still highlight different separabilities.
As for in-differentiable splits, it is near random accuracy, even directly trained on their embeddings.
\item The DB Score curve shows emergent behavior on in-differentiable domains with model size increases.
The PubMed and Pile-CC domains did not show a decreasing DB Score in 410m.
However, when reaching the 2.8b size, their DB scores suddenly decreased in deep layers, meaning that the separability between members and non-members increased, which is even more significant in the 12b model.
This helps to explain the aforementioned RUC-AUC performance boost from 1b to 2.8b since the embeddings of some domains suddenly become more separable in the 2.8b model size, leading to higher RUC-AOC performance.
\item The DB Score bounces back to a high value in the final layer, meaning a decreased separability.
As current MIA methods use the last layer and its computation results (likelihood, tokens, etc.), this may help to explain why MIA performance is low in general, as the last layer itself is not a good option as its embedding separability is low.
%and \cite{men2024shortgptlayerslargelanguage} also suggested the last layer is good at generating tokens, which 
\end{asparaenum}

\subsection{Generation Entropy Dynamics}
\begin{figure}[ht] 
\centering
\includegraphics[width=1\columnwidth, height=0.55\columnwidth]{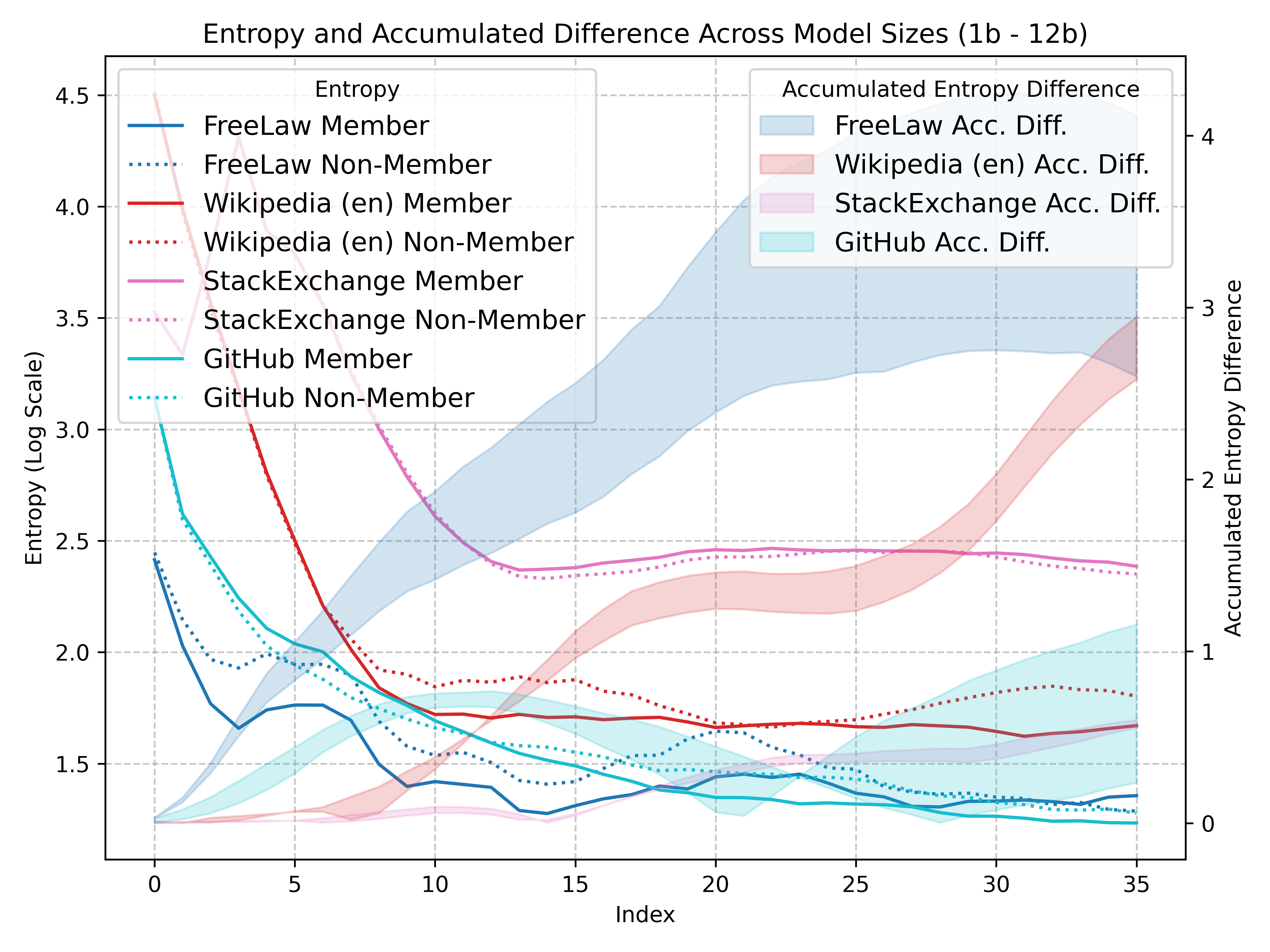} 
\caption{Entropy and accumulated entropy difference over decoding steps in different domains. We only draw average values for entropy to improve readability.}
\label{fig: entropy analysis}
\end{figure}
Current MIA methods pay less attention to the token decoding dynamics in the LLM generation process.
We calculate the token entropy for members and non-members and their accumulated entropy difference across steps in Figure \ref{fig: entropy analysis}.

\begin{asparaenum}[(I)] 
\item From Figure \ref{fig: entropy analysis}, a low or high domain entropy (GitHub, StackExchange) does not relate to its MIA performance in Figure \ref{fig: general results} (c).
However, the domain-dependent entropy (decoding probability) means a domain-dependent log-likelihood, which explains the low threshold generalizability of Gray-Box methods.
This also helps to explain the better performance of Min-k\% ++ as it standardizes the log-likelihood of input tokens, erasing such domain or input text dependency.
\item Though decoding entropy at each step does not show obvious features related to the MIA performance, the accumulated entropy difference increases with the decoding steps, suggesting non-members have a statistically higher entropy compared to the member texts.
Additionally, the domains with higher MIA performance (FreeLaw, Wikipedia (en) in Figure \ref{fig: general results}) have a higher increasing speed in the accumulated entropy difference than the other statistically low MIA performance domains (StackExchange, GitHub).
\end{asparaenum}

\section{Conclusion}
In this study, we revisited the MIA statistically with in-depth analysis from multiple perspectives.
Our results show MIA performance improves with model size and varies across domains, with most MIA methods showing no advantage compared to baselines.
Our results generally support previous negative results, but notable amounts of MIA performance outliers make space for positive results, connecting the MIA performance inconsistency.
We also found that deciding a threshold in MIA is an overlooked challenge.
Additionally, long text and text dissimilarity benefit the MIA performance.
The separability of members and non-members is also reflected in the LLM embedding with emergent change that benefits MIA in large models.
The final layer used by current MIA methods may be a bad choice due to low embedding separability.
Finally,  differentiable members and non-members have faster accumulated entropy difference.

\section{Limitations}
The analysis of the results is mostly based on the statistical level.
This means we do not make assumptions about the correctness of analysis in previous results, and the statistical analysis should be a stand-alone analysis.
The results may not totally align with previous results that were conducted in their own settings.
Additionally, as Pythia only provides model sizes up to 12b, we cannot scale the model size further.
Additionally, only very few LLMs released their pre-train data, and their pre-train data is different, so it is hard to conduct such experiments across models. 
\footnote{Most famous open-source LLMs, like the LLaMA series or Qwen series, did not release their pre-train data.}

Though we tried to extend the scale of the experiments further, the size of the test and valid data limited it.
Their texts will be exhausted with further samplings and no longer satisfy the experiment requirements, where we want sampled members and non-members to be different each time.

We are not able to fully implement all existing MIA methods, but we selected methods that we considered to be representative at the time of this research, with most of those methods published very recently.
There are multiple ways to select a threshold, and there are pros and cons in choosing different calculation methods for a threshold. 
We did not choose to iterate all possible options but chose the geometric balance between TPR and FPR.
We do deny the existence of better threshold calculation method exits that may need different results, but this study is analysis-oriented rather than enumerating possible options to find a better method.

\section{Ethical Considerations}
The original Pile date was reported to contain content related to copyright issues. 
The domains reported with copyright issues are Books3, BookCorpus2, OpenSubtitles, YTSubtitles, and OWT2.
We have made sure we did not conduct any MIA experiment on any of those domains, and we used processed Pile corpora that removed those domains.
This Pile data that removed those domains is accessible online. \footnote{\url{https://huggingface.co/datasets/monology/pile-uncopyrighted}}

For other data we have used, we have made sure the usage aligns with the data license and their intended usage.
Though we conducted experiments over the Pile corpora, we did not observe any personal information or offensive content during the experiments.

\bibliography{anthology,custom}
\bibliographystyle{acl_natbib}

\appendix

\section{Appendix}
\label{sec:appendix}
\subsection{Experiment Setting}
\begin{figure*}[t] 
\centering
\includegraphics[width=1\textwidth]{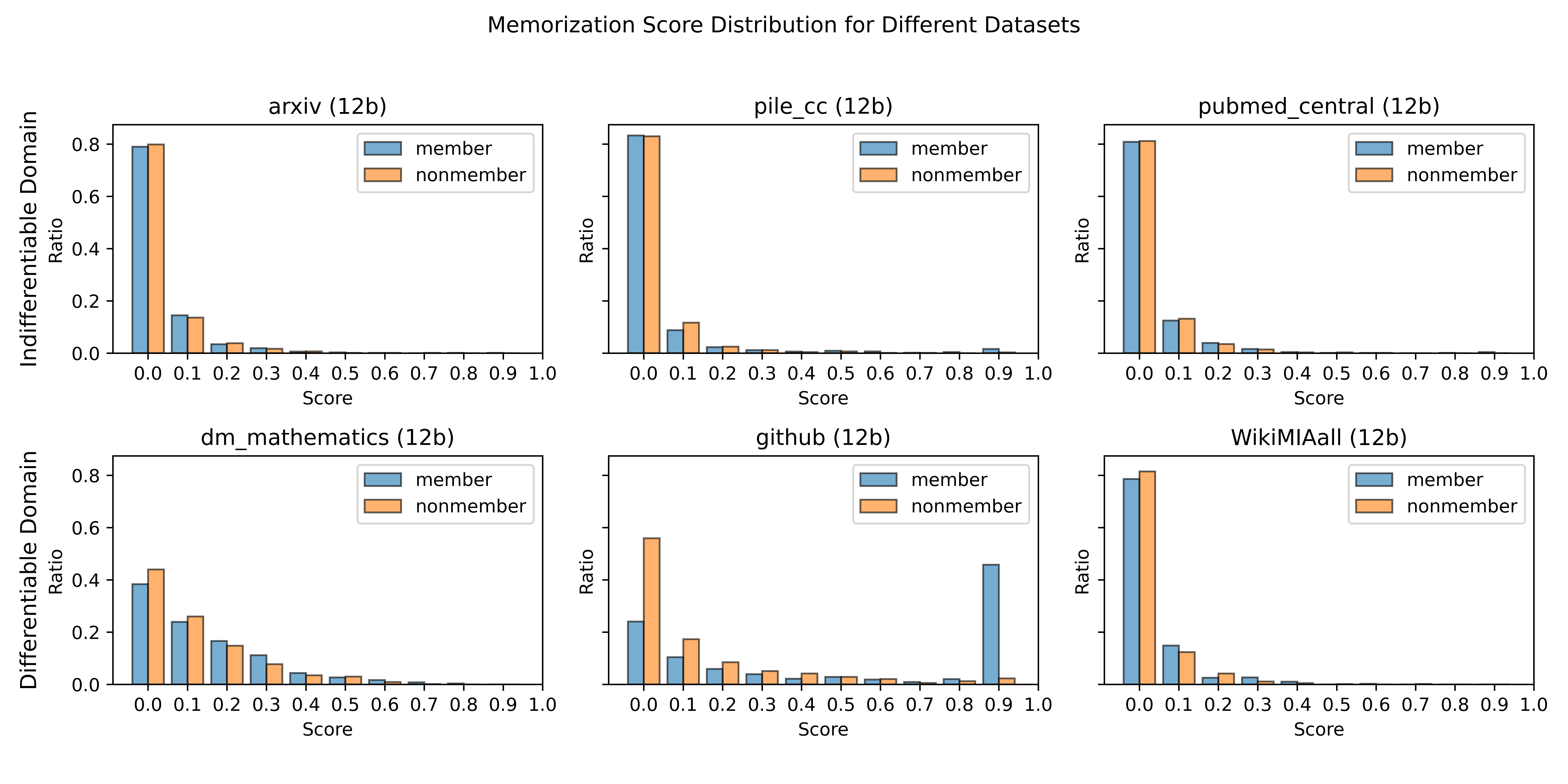} 
\caption{Sampled memorization score distribution in 12b model across domains.}
\label{fig: memorizaiton score sample}
\end{figure*}
The experiment was conducted on over 8 H100 CUDA devices.
Experiments to run one gray-box method overall model sizes take roughly 2 days.
Experiments to run one black-box method over one model size take roughly 20 days, which means the Black-Box methods require more heavy computation as they require the generation of tokens rather than directly taking the intermediate outputs like log-likelihood.
Additionally, unlike Gray-Box methods that can input the entire sentence, the Black-Box requires the input of a partial of this sentence and then requires the LLM to generate the following continuations, in which the generation time will be largely delayed when the input sentence is long.
The random seeds used to sample different member and non-member sets are [47103, 28103, 58320].
For calculating 7-gram overlap, we used Llama 2 tokenizer to tokenize member and non-member texts following to \citet{liu2024infinigramscalingunboundedngram}.
With tokenized texts, we counted the frequencies of appeared 7-gram overlap between member and non-member texts in each split.
We do not use the unique number of 7-gram overlap since we assumed that the frequencies of 7-gram overlap could explain more about text similarity.
However, the frequencies of appeared 7-gram overlap depend on the length of the target texts.
For example, a member and non-member text length of 1,000 has possibly more members than a text length of 100.
Therefore, we normalized it using the sum of the lengths of member and non-member texts to remove the effect brought by the text length.

The Transformer model that is used to predict member and non-member embedding is configured with 256 for its hidden dimension, 2 for the output prediction (member and non-member),  4 for the number of layers, and 8 for the number of attention heads.
It is trained for four epochs with a binary classification objective to classify whether the given input is member text or non-member text.

The entropy is collected by inputting the previous $N$ tokens and asking the LLM to generate the $N+1$ tokens, and then we input the previous $N+1$ tokens.
This process is repeated before reaching the specified text length, which is set as 36 in this experiment.

\subsection{Experiment Setting for MIA Method}
\label{sec: mia setting}
For the reference model, we use the best reference model based on previous research \cite{duan2024membershipinferenceattackswork}.
For the Min-K\% and Min-K\% ++, we choose the K as 20, which means 20\% of $\mathrm{Box}(x)$ are selected from the whole input tokens.
This metric is used in their research paper and repositories.\footnote{https://github.com/zjysteven/mink-plus-plus}\footnote{https://github.com/swj0419/detect-pretrain-code}

For the DC-PDD, there is no hyperparameter, and it relies on a pre-computed token frequency from corpora, which is not released at the time of writing.
To reproduce this study, we used the infini-gram package \footnote{\url{https://infini-gram.io/pkg_doc.html}} as the pre-computed frequency.
However, their frequency is computed over the LLaMa tokenizer, which is different from that of the Pythia tokenizer.
We have to align their results, but this causes inevitable errors, which we cannot manage since the frequency is computed on a different tokenizer, and a sentence may be tokenized into different tokens based on the tokenizer.

For the EDA-PAC, the percentage of words that are swapped is set as 30\%, and collect five perturbed sentences.

For the \textsc{ReCaLL}, the number of shots (the number of prefixes) inserted into the input text is set as 12 while the maximum length is 1,000.
If the prefix combines the input text above the max length of the model, we decrease the number of prompts gradually until the length is acceptable.

For the SaMIA, for a given input prefix, we generate the ten possible continuations with 0.8 temperature, which follows the setting in their original repository. \footnote{https://github.com/nlp-titech/samia}
The model that is used to calculate the semantic similarity is BLUERT-20.\footnote{https://huggingface.co/lucadiliello/BLEURT-20}
As the BLUERT-20 only accepted token lengths up to 512, we are only able to run it up to the length of 500 for this method.

For the CDD, for a given input prefix, we also generate ten possible continuations with 0.8 temperature.

For both SaMIA and CDD methods, the maximum input tokens are 512 based on the input text length, and we generate the rest of the tokens based on the difference with the maximum text length setting in our experiments (1,000 tokens).

\begin{table*}[t]\small
\centering
\begin{tabular}{lcc}
\toprule
Name            & Category & Calculated Feature \\ \midrule
Loss &Gray-Box&  $ \text{Example Loss}$      \\
Perplexity  &  Gray-Box  & \text{Example Perplexity}  \\
Gradient  &  Gray-Box  & \text{Example Gradient}  \\
Reference Model &   Gray-Box       &  $ \text{Loss}_{\text{Target}} - \text{Loss}_{\text{Reference}}$      \\
Zlib Entropy    &   Gray-Box       &   $ \frac{\text{Loss}}{\text{Entropy}(\text{Text})}$   \\
Min-k\% Prob    &    Gray-Box       &  $ \frac{1}{N} \sum_{i=1}^{N} \log P(w_i | w_{<i})$ for $w_i \in \text{Bottom-K\%}$      \\
Min-k\% Prob ++ &     Gray-Box      &   $ \frac{\text{Min-k\% Prob} - \mu}{\sigma}$     \\
DC-PDD             &    Gray-Box      &   Compare the decoded log-likelihood with statistics from large corpora.     \\ 
\textsc{ReCaLL}          &    Gray-Box       &   $\frac{LL(x|p)}{LL(x)}$ when insert $p$ text prefix     \\
SaMIA           &   Black-Box       &   $\frac{1}{N} \sum_{i=1}^{N} \text{SemanticDistance}(g_i, a)$ .     \\
CDD             &    Black-Box      &   $ \frac{1}{N} \sum_{i=1}^{N}\text{EditDistance}(g_i, a)$.     \\ 

\bottomrule
\end{tabular}
\caption{Collection of MIA methods evaluated in this study.}
\label{tab:method table}
\end{table*}

\subsection{Available Domains in Each Split Method}
We have the Truncate, Complete, and Relative split method over the input text of all domains in the Pile corpora.
We only keep those splits that have at least 100 examples for both member and non-member text at all text lengths.
If a domain does not meet this requirement, it will be discarded.
The available domains for all those split methods are presented in the following Table \ref{tab:shared_domains} and \ref{tab:all_domains}.

For each MIA method, the results run on all of its split methods, length range, model size, and random seeds.
We are also able to see that the Complete splitting methods have the lease domains as the whole length of a text is a strict standard.
Additionally, we also see that Relative split has the most domains as this split method suits the distribution of the target domains. Thus, most data are kept using this split method while following the text distribution.

\begin{table*}[t]
\centering

% Subtable 1
\begin{subtable}{\textwidth}
\centering
\caption{Domains in shared split methods.}
\begin{tabular}{lc}
\toprule
\textbf{Split Method} & \textbf{Shared Domains} \\ \midrule
Truncated & Wikipedia (en), StackExchange, Pile-CC, GitHub, FreeLaw \\
Complete & Wikipedia (en), StackExchange, Pile-CC, GitHub, FreeLaw \\
Relative & Wikipedia (en), StackExchange, Pile-CC, GitHub, FreeLaw \\
\bottomrule
\end{tabular}
\label{tab:shared_domains}
\end{subtable}

\vspace{0.5cm} % Small vertical space between subtables

% Subtable 2
\begin{subtable}{\textwidth}
\centering
\caption{Domains in specific split methods.}
\begin{tabular}{lc}
\toprule
\textbf{Split Method} & \textbf{Specific Domains} \\ \midrule
Truncated & PubMed Central, HackNews, EuroParl, DM Mathematics, arXiv \\
Complete & USPTO Backgrounds \\
Relative & PubMed Central, NIH ExPorter, HackNews, Enron Mails, DM Mathematics, arXiv \\
\bottomrule
\end{tabular}
\label{tab:all_domains}
\end{subtable}

\caption{Domains included in different split methods across shared and specific datasets.}
\label{tab:split_domains}
\end{table*}

\subsection{Memorization and MIA}

\subsubsection{Memorization Score Sample Distribution}

In this figure, we saw that LLM does not show a very obvious distribution gap for most of the domains.
However, we notice that in the GitHub domain, there are many texts that show high memorization scores, meaning that most of the texts are well-memorized by the LLM.
A similar trend is also observed in DM Mathematics, where most of the texts are not distributed in the low memorization score area (memorization score 0 - 0.2).
A similarity between those two domains is that both of those two domains have low vocabulary diversity, considering that both DM Mathematics and GitHub are more oriented in symbols (math) or fixed expression (coding). Thus, it is easier for the LLM to memorize those texts.

\begin{figure}[t] 
\centering
\includegraphics[width=1\columnwidth]{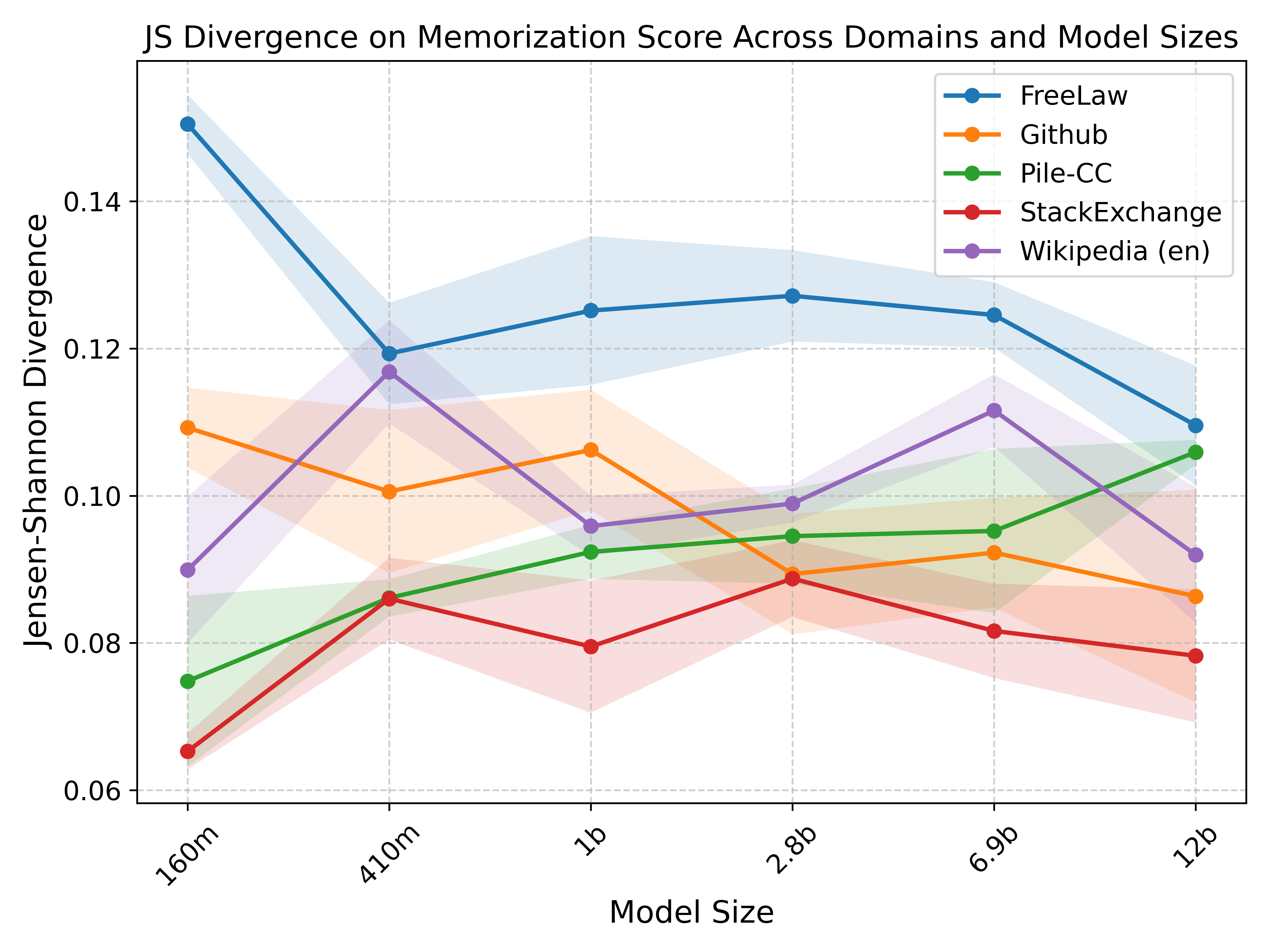} 
\caption{Memorization Score Distribution Divergence for Member and Non-Member Text}
\label{fig: memorization score analysis}
\end{figure}
% The black-box MIA methods aim to compare the similarity between generated and actual text, expecting higher similarity for member texts.
% However, LLMs can generate text that matches actual text, even for non-member texts, leading to a false sense of membership inference. 
\subsubsection{Memorization Score Distribution Distance}

This section examines whether MIA performance is related to memorization by comparing the generated tokens with actual continuations when promoting 32 tokens, known as the K-extractable score, for both non-member and member text. 
We compute the distribution distance between the K-extractable score over the different domains on member and non-member text using JS divergence, as shown in Figure \ref{fig: memorization score analysis}.
We can see a correlation between MIA performance and the JS distribution difference between member and non-member text. 
The FreeLaw has the highest JS Divergence score among those domains, suggesting that the memorization score distribution between member text and non-member text is large. 
This aligns with the ROC-AUC score density distribution in Figure \ref{fig: general results}.
Additionally, we also see that GitHub and StackExchange have a low divergence, meaning their memorization score distribution between member text and non-member text is small, which is hard to differentiate.
\subsection{Membership Inference Attack as Hypothesis Test}
Besides directly analyzing the probability density function of the ROC-AUC scores, we also try to look at the MIA from a hypothesis test perspective.
We treat the feature scores of member and non-member text as two distributions and use a hypothesis test to verify whether those two distributions are the same distributions or not.
If a distribution passes such verification, it at least means the feature score distribution of the member is different from the feature distribution of the non-member text.
Even though it does not guarantee any MIA performance, it does not directly evaluate MIA performance; it just shows whether those two distributions are the same or not.
Such analysis at least provides a perspective to look at MIA differently.
We divide the number of splits whose feature scores of member and non-member passed the verification as two distributions with the total amounts of splits.
The results are presented from Table \ref{tab:hp1} to \ref{tab:hp3}.
\begin{asparaenum}
    \item Similar to MIA performance, we observed that the number of splits that pass the hypothesis test increases with the model size. 
    This confirms the analysis of the results of the RUC-AOC score using the probability density functions.
    \item Further, we also see that in this evaluation metrics, the best-performed method Min-K\% ++ does not also show the best performances in passing the hypothesis test.
    On the contrary, the best-performed MIA method is the Refer, which actually has the lowest performance in the ROC-AUC analysis.
    The reason is that the hypothesis test method does not evaluate whether the two examples are separate or not; it evaluates how those two distributions consisting of the members and non-members are the same distribution or not.
    This means that they do not consider separating a specific example but focus on identifying those two distributions.
    \item Even though the hypothesis test does not provide a method to differentiate members and non-members specifically.
    It tells the performance of MIA from another perspective, whereas the previous worst-performing method could actually have the best performance.
    This shows the importance of evaluating the MIA method from multiple perspectives rather than only focusing on certain metrics, which could be misleading. 
    \item In this metric, we are also able to observe the same performance boost when transferring from the 1b to 2.8b model.
    This aligns with the observation in the probability density analysis of RUC-AOC scores across dimensions, which confirms the emergent embedding change that we have discovered.    
\end{asparaenum}

\begin{table}[t]\scriptsize
\centering
\begin{tabular}{lcccccc}
\toprule
\textbf{Method} & \textbf{160m} & \textbf{410m} & \textbf{1b} & \textbf{2.8b} & \textbf{6.9b} & \textbf{12b} \\ \midrule
Loss & 0.08 & 0.067 & 0.087 & 0.107 & 0.120 & 0.167 \\ 
Min-K\% & 0.08 & 0.087 & 0.073 & 0.153 & 0.207 & 0.220 \\ 
Zlib & 0.013 & 0.020 & 0.040 & 0.080 & 0.113 & 0.173 \\ 
SaMIA & 0.093 & 0.073 & 0.053 & 0.080 & 0.060 & 0.080 \\ 
Min-K\% ++ & 0.033 & 0.073 & 0.133 & 0.273 & 0.453 & 0.567 \\ 
Refer & 0.040 & 0.040 & 0.113 & 0.533 & 0.740 & 0.787 \\ 
Grad & 0.033 & 0.033 & 0.053 & 0.133 & 0.160 & 0.047 \\ 
DC-PDD & 0.073 & 0.080 & 0.087 & 0.160 & 0.140 & 0.240 \\ 
CDD & 0.033 & 0.060 & 0.093 & 0.027 & 0.087 & 0.060 \\ 
\textsc{ReCaLL} & 0.093 & 0.067 & 0.107 & 0.140 & 0.140 & 0.167 \\ \bottomrule
\end{tabular}
\caption{Hypothesis test results across MIA methods and model size in Relative Split method.}
\label{tab:hp1}
\end{table}

\begin{table}[!tb]\scriptsize
\centering
\begin{tabular}{lcccccc}
\toprule
\textbf{Method} & \textbf{160m} & \textbf{410m} & \textbf{1b} & \textbf{2.8b} & \textbf{6.9b} & \textbf{12b} \\ \midrule
Loss & 0.033 & 0.053 & 0.080 & 0.120 & 0.140 & 0.193 \\ 
Min-K & 0.033 & 0.073 & 0.113 & 0.187 & 0.227 & 0.313 \\ 
Zlib & 0.040 & 0.027 & 0.040 & 0.107 & 0.167 & 0.260 \\ 
SaMIA & 0.180 & 0.153 & 0.160 & 0.153 & 0.140 & 0.140 \\ 
Min-K\% ++ & 0.053 & 0.027 & 0.093 & 0.293 & 0.460 & 0.553 \\ 
Refer & 0.027 & 0.020 & 0.147 & 0.527 & 0.733 & 0.760 \\ 
Grad & 0.020 & 0.060 & 0.047 & 0.073 & 0.067 & 0.060 \\ 
DC-PDD & 0.060 & 0.073 & 0.100 & 0.187 & 0.240 & 0.360 \\ 
CDD & 0.033 & 0.040 & 0.053 & 0.047 & 0.080 & 0.040 \\ 
EDA-PAC & 0.033 & 0.080 & 0.060 & 0.047 & 0.060 & 0.047 \\ 
\textsc{ReCaLL} & 0.053 & 0.060 & 0.087 & 0.127 & 0.133 & 0.207 \\ \bottomrule
\end{tabular}
\caption{Hypothesis test results across MIA methods and model size in Truncate Split method.}
\label{tab:hp2}
\end{table}

\begin{table}[!tb]\scriptsize
\centering
\begin{tabular}{lcccccc}
\toprule
\textbf{Method} & \textbf{160m} & \textbf{410m} & \textbf{1b} & \textbf{2.8b} & \textbf{6.9b} & \textbf{12b} \\ \midrule
Loss & 0.053 & 0.033 & 0.073 & 0.127 & 0.100 & 0.113 \\ 
Min-K & 0.087 & 0.080 & 0.080 & 0.153 & 0.147 & 0.247 \\ 
Zlib & 0.100 & 0.060 & 0.060 & 0.147 & 0.147 & 0.200 \\ 
SaMIA & 0.120 & 0.153 & 0.153 & 0.113 & 0.120 & 0.153 \\ 
Min-K\% ++ & 0.053 & 0.087 & 0.133 & 0.387 & 0.413 & 0.547 \\ 
Refer & 0.040 & 0.040 & 0.107 & 0.500 & 0.607 & 0.633 \\ 
Grad & 0.027 & 0.020 & 0.027 & 0.193 & 0.147 & 0.053 \\ 
DC-PDD & 0.067 & 0.073 & 0.073 & 0.200 & 0.213 & 0.300 \\ 
CDD & 0.040 & 0.093 & 0.060 & 0.060 & 0.013 & 0.060 \\ 
EDA-PAC & 0.067 & 0.060 & 0.047 & 0.067 & 0.033 & 0.027 \\ 
\textsc{ReCaLL} & 0.053 & 0.033 & 0.073 & 0.127 & 0.100 & 0.113 \\ \bottomrule
\end{tabular}
\caption{Hypothesis test results across MIA methods and model size in Complete Split method.}
\label{tab:hp3}
\end{table}
\begin{figure*}[t] 
\centering
\includegraphics[width=1\textwidth, height=0.5\textwidth]{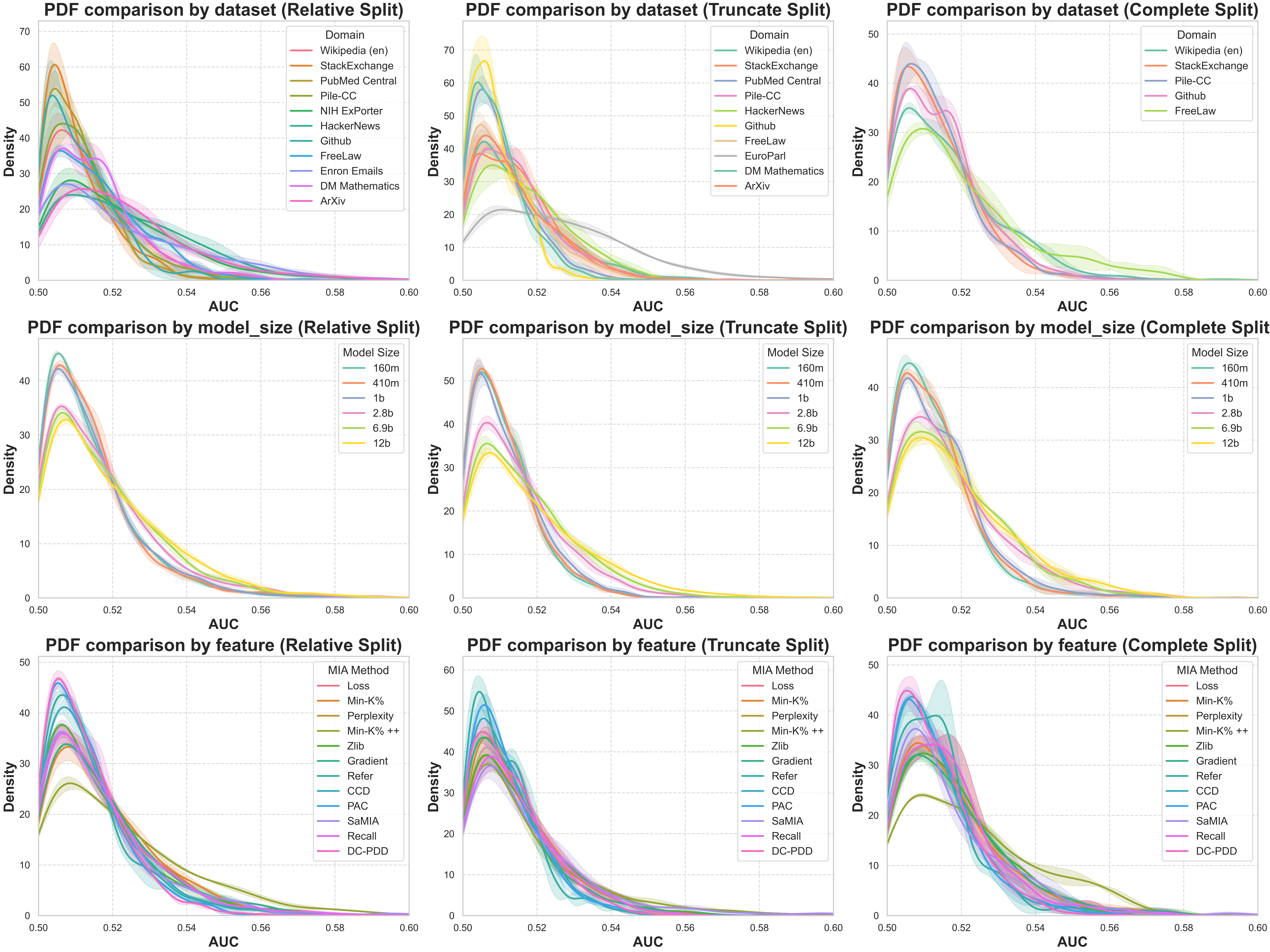} 
\caption{Detailed Results in Each Split Method}
\label{fig: all category results}
\end{figure*}
\subsection{Detailed Results in Each Split Method}
\label{other pdfs}
In this section, we present the detailed results for Truncate Split, Complete Split, and Relative Split.
Each split contains all available domains.
We shot the probability density in the Domain, Model Size, and MIA Method dimension in Figure \ref{fig: all category results}.
\begin{asparaenum}
    \item In the first row, which shows the probability density over domains, we saw some more high-performance domains.
    For example, in the Truncate split, the EuroParl performs very well compared to other domains.
    One of the reasons may be that the EuroParl contains some non-English texts, which serve as an important feature for the member and non-member classification.
    Still, in the relative split, we are able to see more domains with relatively high MIA performance compared to other domains, which helps to explain why Relative Split can give better performance.
    \item In the second row, which shows the probability density over model sizes, we saw a uniform performance across different splits where the MIA performance positively scales with the model size.
    \item In the third row, which shows the probability density over the different MIA methods. We are also able to observe some split-based differences.
    In the Relative and Complete split, we can see that the Min-k\% ++ performs better than other methods.
    However, in the Truncate split, we see mixed results where most methods do not show obvious performance differences, where the Min-k\% ++ is no longer significantly better than other methods.
\end{asparaenum}
\begin{figure*}[htbp] 
\centering
\includegraphics[width=1\textwidth, height=0.35\textwidth]{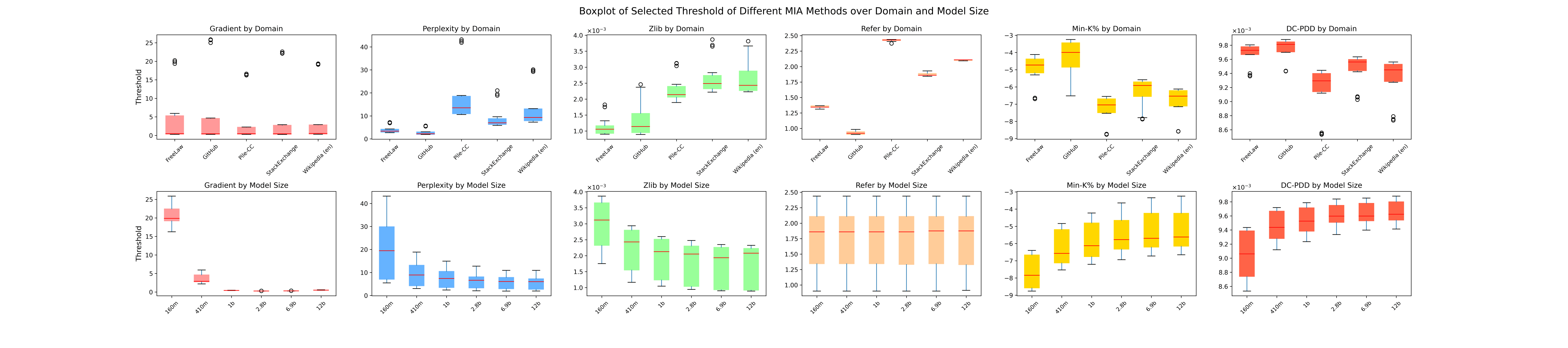} 
\caption{Boxplot of other MIA methods across model sizes and domains}
\label{fig: box plot other}
\end{figure*}
\section{Boxplot of the threshold for other MIA methods}
\label{other boxplots}
In this section, we show the boxplot of the threshold for other MIA methods across model sizes and domains in Figure \ref{fig: box plot other}.
From this figure, we can obtain the following:

\begin{asparaenum}
    \item The threshold still changes across domains with the existence of outliers for all those methods. The Refer method shows an extreme trend where the threshold in each domain is totally different, indicating a threshold decided from another domain totally failed to generalize to other domains.
    \item Regarding the model size, we still observe that their thresholds change across model sizes.
    However, the Refer model has a stable threshold that is generalized well in other model sizes.
    This is probably because the referee relies on a reference model, which makes it less dependent on the target model.
    However, it still contains outliers, indicating that the threshold in one model size may not work in another model size depending on the input members and non-members.
    \item  Additionally, while the trend is not general, we are able to see that the change of threshold is not random in some methods. 
    We saw that the perplexity, zlib, Min-k\%, and DC-PDD all showed either a gradual increase or decreasing threshold values.
    This can increase the predictability of the threshold, making the decision of the threshold less random.
    However, even if it indicates some trend, it is still hard to make a correct prediction regarding how the threshold would change across model sizes.
\end{asparaenum}

\end{document}